\documentclass[runningheads]{llncs}
\usepackage{graphicx}
\usepackage{graphicx}
\usepackage{hyperref}

\usepackage{multirow}
\usepackage{tabu}
\usepackage{bbm}
\usepackage{diagbox}
\usepackage[english]{babel}
\usepackage{comment}
\usepackage{array}
\usepackage{amstext}
\usepackage{makecell}
\usepackage{caption}
\usepackage{tablefootnote}
\usepackage{babel}
\usepackage{booktabs} 
\usepackage{color}
\usepackage{amsmath}
\usepackage{wrapfig}
\usepackage{amssymb}

\newif\ifdraft
\draftfalse
\drafttrue

\ifdraft
 \newcommand{\PF}[1]{{\color{red}{\bf PF: #1}}}
 
 \newcommand{\MS}[1]{{\color{green}{\bf MS: #1}}}
 
 \newcommand{\ZD}[1]{{\color{violet}{\bf ZD: #1}}}
 
 \newcommand{\YH}[1]{{\color{blue}{\bf YH: #1}}}
 
 \newcommand{\SPE}[1]{{\color{orange}{\bf SS: #1}}}
 
 \newcommand{\WJ}[1]{{\color{orange}{\bf WJ: #1}}}
 
  \newcommand{\JY}[1]{{\color{blue}{\bf JY: #1}}}
 
  \newcommand{\placeholder}[1]{{\color{green}{placeholder: #1}}}
\else
 \newcommand{\PF}[1]{}
 
 \newcommand{\KY}[1]{}
 
 \newcommand{\MS}[1]{}
 
 \newcommand{\ZD}[1]{}
 
 \newcommand{\YH}[1]{}
 
 \newcommand{\SPE}[1]{}
 
 \newcommand{\WJ}[1]{}
 
 \newcommand{\JY}[1]{}
 
 \newcommand{\placeholder}[1]{}
\fi

\newcommand{\bp}{\mathbf{p}}
\newcommand{\bz}{\mathbf{z}}

\newcommand{\cF}{\mathcal{F}}

\newcommand{\ie}{\textit{i.e.}}
\newcommand{\eg}{\textit{e.g.}}

\begin{document}

\title{Neural Annotation Refinement: Development of a New 3D Dataset for Adrenal Gland Analysis}

\titlerunning{Neural Annotation Refinement for ALAN Dataset Development}
\author{Jiancheng Yang\inst{1,2,}\thanks{These authors have contributed equally: Jiancheng Yang and Rui Shi.}  \and Rui Shi\inst{1,\star} \and Udaranga Wickramasinghe\inst{2} \and \\Qikui Zhu\inst{3,4} \and Bingbing Ni\inst{1}\thanks{Corresponding author: Bingbing Ni (nibingbing@sjtu.edu.cn).} \and Pascal Fua\inst{2}}

\authorrunning{J. Yang et al.}

\institute{Shanghai Jiao Tong University, Shanghai, China\\
\email{nibingbing@sjtu.edu.cn}	\\
	\and EPFL, Lausanne, Switzerland
	\and Dept. of Computer and Data Science, Case Western Reserve University, OH, USA
	\and Dept. of Biomedical Engineering, Case Western Reserve University, OH, USA
}
\maketitle 
%


\begin{abstract}

The human annotations are imperfect, especially when produced by junior practitioners. Multi-expert consensus is usually regarded as golden standard, while this annotation protocol is too expensive to implement in many real-world projects. In this study, we propose a method to refine human annotation, named \textit{Neural Annotation Refinement (NeAR)}. It is based on a learnable implicit function, which decodes a latent vector into represented shape. By integrating the appearance as an input of implicit functions, the appearance-aware NeAR fixes the annotation artefacts. Our method is demonstrated on the application of adrenal gland analysis. We first show that the NeAR can repair distorted golden standards on a public adrenal gland segmentation dataset. Besides, we develop a new Adrenal gLand ANalysis (ALAN) dataset with the proposed NeAR, where each case consists of a 3D shape of adrenal gland and its diagnosis label (normal vs. abnormal) assigned by experts. We show that models trained on the shapes repaired by the NeAR can diagnose adrenal glands better than the original ones. The ALAN dataset will be open-source, with 1,584 shapes for adrenal gland diagnosis, which serves as a new benchmark for medical shape analysis. Code and dataset are available at \url{https://github.com/M3DV/NeAR}.

\end{abstract}

\section{Introduction}

\begin{figure}[!tb]
    \centering
	\includegraphics[width=0.8\linewidth]{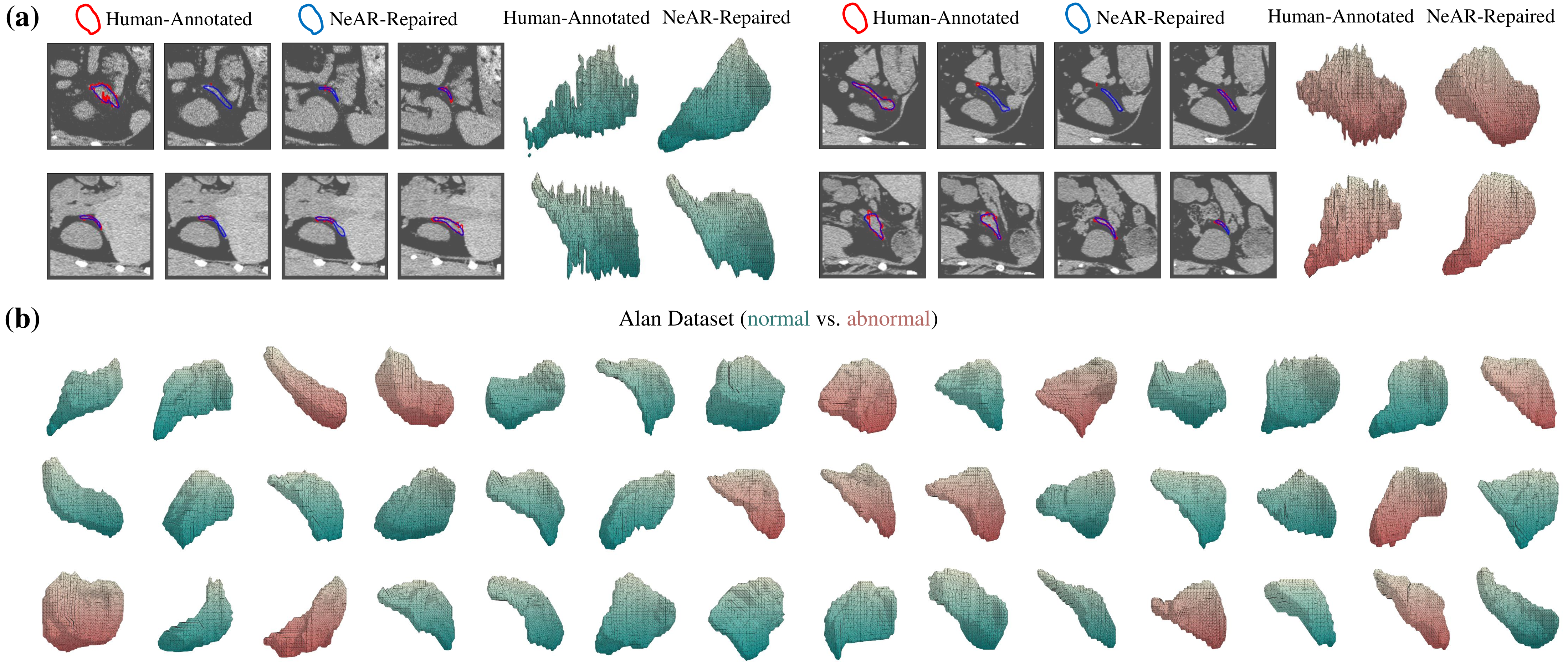}
	\caption{\textbf{The ALAN Dataset.} It features 1,584 adrenal glands, each one of which has been tagged by human experts as normal or abnormal. The normal ones are shown in green and the abnormal in red. Best viewed on screen. \textbf{(a)} Images with human and NeAR-repaired annotations, in red and blue contours respectively, and the corresponding 3D visualization.  \textbf{(b)} The repaired ALAN Dataset.	}
	\label{fig:alan_dataset}
\end{figure}

Deep learning has enjoyed a great success in medical image analysis, but large annotated datasets are required to achieve this~\cite{gulshan2016development,esteva2017dermatologist,kermany2018identifying,ardila2019end,esteva2021deep}.
Unfortunately, such datasets are difficult to obtain in part because human annotations are known to be imperfect~\cite{karimi2020deep,tajbakhsh2020embracing}.
In medical image segmentation, multi-expert consensus is employed as golden standard, where the agreement of multiple annotators is regarded as ground truth. Nevertheless, this protocol involving multiple medical experts is often too time-consuming and expensive to achieve in practice. In 
those cases, there are often high-frequency artefacts and false positive/negative in human segmentation. Please refer to Fig.~\ref{fig:alan_dataset}~(a) for illustration.


In this paper, we introduce  \emph{Neural Annotation Refinement (NeAR)}, an approach to automatically correcting human-annotated segmentation databases, so that networks trained using the corrected database perform better than those trained using the original one. Our method is developed based on the fact that a neural network with appropriate inductive bias could serve as a deep prior~\cite{ulyanov2018deep,hanocka2020point2mesh}. By leveraging the recent advance in implicit surface modeling~\cite{chen2019learning,mescheder2019occupancy,park2019deepsdf} that uses a neural network (\eg{}, MLP and CNN~\cite{peng2020convolutional}) as 
a mapping from spatial coordinates to a shape representation, the NeAR learns data-efficient implicit functions as a shape prior of the target annotations, which can be used to repair human annotations. To make the repaired segmentation appearance-aware, we integrate the appearance as an input of the implicit function. As illustrated in Fig.~\ref{fig:alan_dataset}~(a), the repaired segmentation by the proposed NeAR is visually appealing. We will further show that the repairing can be used to improve downstream applications. 

Our method is demonstrated on the application of adrenal gland analysis. We first show that the NeAR can repair distorted golden standards on a public adrenal gland segmentation dataset, consisting of 100 cases. The NeAR outperforms standard segmentation methods quantitatively in terms of the repaired annotation quality. Furthermore, we apply the NeAR to repair a new Adrenal gLand ANalysis (ALAN) dataset of 1,584 cases, where each adrenal gland is segmented by 1 clinician and diagnosed---as normal or abnormal---by 2 clinicians and 1 senior endocrinologist. In other words, the diagnosis label is quite reliable whereas the segmentation exhibits problems as shown in Fig.~\ref{fig:alan_dataset}~(a).  As shown in Fig.~\ref{fig:alan_dataset}~(b), NeAR can effectively repair these segmentations, as evidenced by the fact that models trained on the shapes repaired by the NeAR can better diagnose adrenal glands (normal vs. abnormal) than the original ones. 

As an independent contribution, the ALAN dataset will be open-source, with NeAR-repaired shapes of adrenal glands and the corresponding diagnosis labels (normal vs. abnormal), as illustrated in Fig.~\ref{fig:alan_dataset}~(b). This shape classification benchmark with 1,584 high-quality 3D models will be of interest for medical image analysis and geometric deep learning research community.\footnote[1]{Code and dataset are available at \url{https://github.com/M3DV/NeAR}.}


\section{Method}

In this section, we first briefly review deep implicit surface, an emerging technique in 3D vision. We then introduce how this technique can be applied to repair human annotated segmentation labels, and propose the Neural Annotation Refinement (NeAR) based on appearance-aware implicit surface model.

\subsection{Deep Implicit Surfaces}

Implicit surface modeling~\cite{chen2019learning,mescheder2019occupancy,park2019deepsdf} maps spatial coordinates to shape representations with a neural network. Typically, the shape representation could be either binary occupancy or signed / unsigned distance. For simplicity, we use occupancy fields~\cite{mescheder2019occupancy} in this study, while the whole framework can be easily applied on distance functions~\cite{park2019deepsdf}. In implicit surface modeling, a 3D shape is first encoded with a $c$-dimensional latent vector $\bz\in \mathbb{R}^c$, and a continuous representation of the shape is then obtained by learning a mapping: 
\begin{equation}
    \cF(\bz, \bp)=o:\mathbb{R}^c\times \mathbb{R}^3 \rightarrow [0,1].
\end{equation}
Here, a $c$-dimensional latent vector $\bz\in \mathbb{R}^c$ and coordinates of a query point $\bp\in \mathbb{R}^3$ are inputted into a neural network $\cF$--typically multi-layer perceptron (MLP)--to classify whether the query point is inside or outside the represented shape, with the occupancy probability $o$ close to $1$ for $\bp$ inside the shape and $0$ otherwise. With a thresholding parameter $t$, the underlying surface is implicitly represented by the decision boundary $\cF(\bz,\bp)=t$.
For model training, we apply auto-decoding~\cite{park2019deepsdf}, an encoder-free approach where a learnable latent vector $\bz$ of each shape is directly taken as input, jointly optimized with the parameters of $\cF$ through back-propagation. The number of latent vectors is equal to the number of training shape samples. 

The deep implicit models have achieved a great success in a wide range of applications, \eg{}, shape modeling~\cite{chen2019learning,mescheder2019occupancy,park2019deepsdf,huang2022representation}, 3D reconstruction~\cite{chibane2020implicit,xu2019disn} and differentiable rendering~\cite{mildenhall2020nerf,niemeyer2021giraffe}. The deep implicit surface serves as a deep prior~\cite{ulyanov2018deep,hanocka2020point2mesh} for shape modeling, thus can be used as a tool to refine annotations, as the implicit reconstructions tend to remove high-frequency artefacts introduced by human annotators. However, standard implicit surface methods are not aware of the appearance, thus the reconstructed surfaces could be misaligned with the actual boundaries. It motivates us to propose the appearance-aware implicit surface model for annotation refinement. Moreover, as the MLP-based implicit functions tend to be data-hungry, which is hard to be satisfied in medical imaging scenario, we introduce the convolutional architecture with multi-scale features to reconstruct the shapes.

\subsection{Neural Annotation Refinement}
\label{sec:method-near}

\paragraph{Appearance-Aware Annotation Refinement.}
The standard deep implicit surface takes spatial coordinates as input; Although the learned shape prior is able to reconstruct a high-quality surface, the reconstructed surface is possible to be misaligned with the actual boundaries. To make the implicit model appearance-aware, we employ a simple strategy by changing the input of the implicit function from spatial coordinates $\bp$ to $\bp$ with its appearance $a$, \ie{},
\begin{equation}
    \cF(\bz, \bp, a)=o:\mathbb{R}^c\times \mathbb{R}^3\times\mathbb{R} \rightarrow [0,1],
\end{equation}
where $a$ short for $a(\bp)$ denotes the image appearance at the position $\bp$, \eg{}, Hounsfield Units in computed tomography. As will be shown, this simple modification leads to significant improvement over shape-only implicit models in both annotation refinement and downstream applications.


\begin{figure}[tb]
    \centering
	\includegraphics[width=0.8\linewidth]{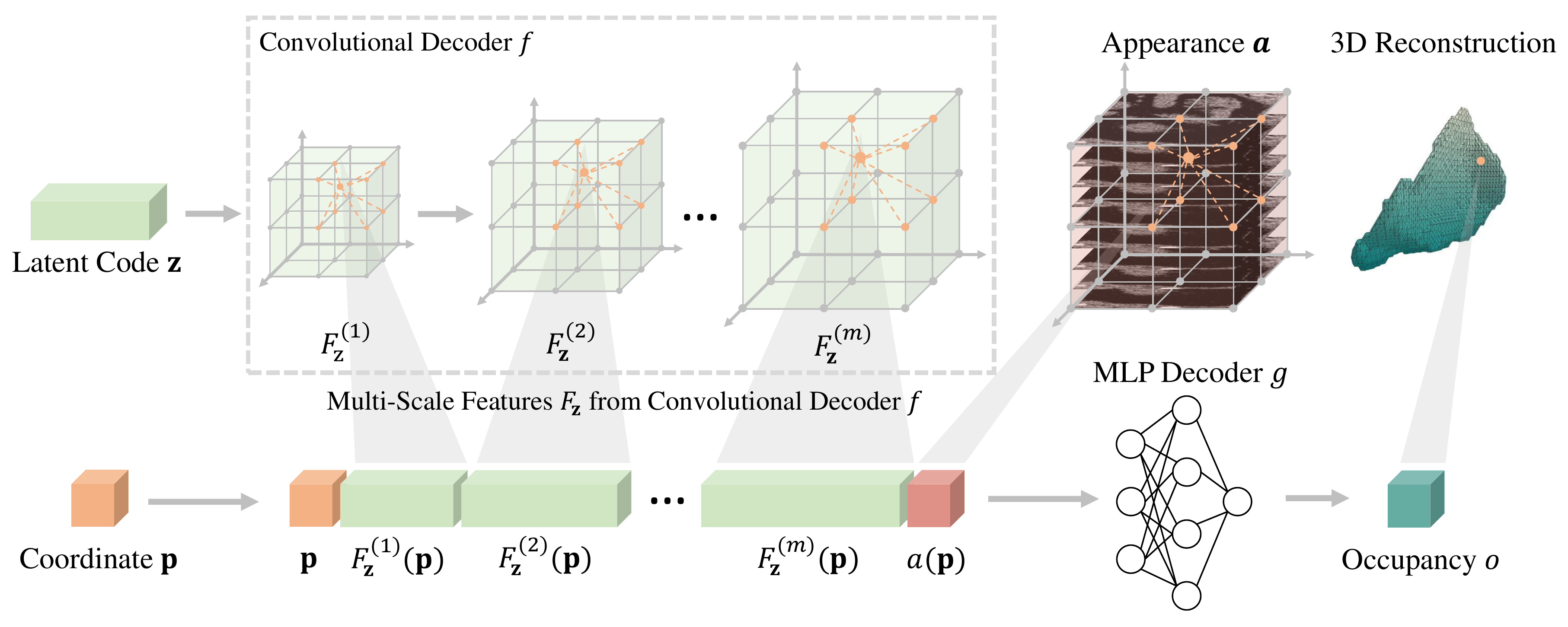}
	\caption{\textbf{Neural Annotation Refinement (NeAR).}  Given a learnable latent vector, it builds multi-scale feature maps $F_\bz$ by a convolutional decoder $f$. A query coordinate $\bp$ aggregates global and local features $F_{\bz}^{(1)},F_{\bz}^{(2)},...,F_{\bz}^{(m)}$, with its appearance $a$ from image. Finally, these point-wise features are fed into a light MLP $g$ for occupancy prediction $o$ to reconstruct appearance-aware surface.}
	\label{fig:near_method}
\end{figure}

\paragraph{Network Architecture.} 
As the data size is generally small in medical imaging applications, standard MLP-based implicit functions can be hard to train. To improve the data efficiency of MLP-based implicit functions, we introduce a convolutional decoder with multi-scale feature aggregation into the deep pipeline, inspired by~\cite{peng2020convolutional,chibane2020implicit,yang2022implicitatlas}. As illustrated in Fig.~\ref{fig:near_method}, a latent vector $\bz$ is first transformed by a convolutional decoder $f$ into multi-scale feature maps, 
\begin{equation}
    F_\bz=[F_\bz^{(1)}, F_\bz^{(2)},\cdots, F_\bz^{(m)}]=f(\bz).
\end{equation}
In our experiments, the resolution of largest feature map $F_\bz^{(m)}$ is $32\times32\times32$.

For a query point $\bp$, it obtains its features $F_\bz^{(1)}(\bp), F_\bz^{(2)}(\bp),\cdots, F_\bz^{(m)}(\bp)$ by trilinear interpolation from the multi-scale maps. 
To make the model appearance-aware, we further integrate the appearance $a=a(\bp)$ as an input of the implicit function. To be concrete, the coordinates and the point-wise features, are concatenated and transformed into the occupancy $o$ through a light MLP $g$, 
\begin{equation}
    o=\cF(\bz,\bp,a)=g(\bp,F_\bz^{(1)}(\bp), F_\bz^{(2)}(\bp),\cdots, F_\bz^{(m)}(\bp),a(\bp)).
\end{equation}

\paragraph{Model Training and Inference.}

We treat the shape implicitly as occupancy field and train the model via auto-decoding~\cite{park2019deepsdf}. The shape loss is measured by binary cross entropy ($\mathit{BCE}$) between predicted occupancy $o$ and ground-truth occupancy $\hat{o}$. Different from standard auto-encoding technique, the auto-decoding is encoder-free. We thus add regularization loss, the $l_2$-norm of the latent code. In total, the training loss is weighted by $\lambda$ ( $\lambda=0.01$ in our experiments),
\begin{equation}
    \mathcal{L}=\mathit{BCE}(o, \hat{o})+\lambda\cdot||\bz||_2.
\end{equation}

The flexibility of implicit functions enables different training resolution from actual resolution. At training stage, we sample $64\times 64\times 64$ meshgrid coordinates from full-resolution $128\times 128\times 128$ input volumes to reduce training cost. The training meshgrid is added by a random Gaussian noise $\mathcal{N}(0,0.01^2)$, whose occupancy labels are sampled from the full-resolution ground truth. We utilize an Adam optimizer~\cite{kingma2014adam} with an initial learning rate of $0.001$ and train the model for $1,500$ epochs. At inference stage, we sample full-resolution uniform meshgrid to reconstruct surfaces, \ie{}, repaired annotations in this study.

\paragraph{Counterpart.}
Segmentation models~\cite{rajchl2016deepcut} with image as input to refine the source masks, \eg{}, 3D ResNet-based FCN~\cite{long2015fully,he2016deep} (Seg-FCN) and 3D UNet~\cite{cciccek20163d} (Seg-UNet), are used as counterparts. They are trained with human-annotated segmentation masks, and try to output refined annotations. Note that the label refinement counterparts are trained with the same datasets.

\section{Datasets}

\subsection{Distorting a Golden Standard Segmentation Dataset}
\label{sec:distorting_golden_standard}

In order to quantitatively analyze the performance of annotation refinement, we synthesize distorted segmentation masks from golden standard. Here, we use the public AbdomenCT-1K~\cite{Ma-2021-AbdomenCT-1K}, an abdominal CT organ segmentation dataset, which is annotated under multi-expert consensus protocol and thus can be regarded as golden standard. We use the adrenal gland subset\footnote[2]{\url{https://github.com/JunMa11/AbdomenCT-1K}}, containing 100 adrenal glands from 50 patients. For each case, we calculate the center of left and right adrenal gland respectively, and center-crop the left and right adrenal gland into $128 \times 128 \times 128$ volumes with a normalized spacing of $1mm^3$. The resulting dataset has 100 cases with golden standard segmentation of adrenal glands. For image pre-processing, we clip the Hounsfield Units using soft-organ window [-60, 140] and then normalize to [0,1].

To synthesize distorted segmentation masks, we randomly add or cut out cubes on the boundary. We then apply random dilation or erosion operation to the shape followed by adding small salt-and-pepper noises. The resulting distorted masks are demonstrated in Fig.~\ref{fig:visualization_golden_standard}, which imitate imperfect human annotations, including high-frequency artefacts (unsmooth boundaries) and false positive/negative. The average Dice between distorted masks and ground truth is $0.71$, with a lower bound of $0.65$ and an upper bound of $0.75$. 

\subsection{ALAN Dataset: A New 3D Dataset for Adrenal Gland Analysis}
\label{sec:alan_dataset}

In this study, we introduce a new 3D Adrenal gLand ANalysis dataset, named ALAN. It consists of computed tomography (CT) scans from 792 patients (\ie{}, 1,584 left and right adrenal glands). Each case is annotated with a segmentation mask and a binary diagnosis label (normal vs. abnormal)~\cite{yang2021medmnist}. The segmentation mask is annotated by a single clinician using 3D Slicer software. As the boundary of adrenal gland--soft organ--is difficult to identify, and the segmentation is made slice-by-slice, the resulting 3D segmentation is imperfect with potential errors, \eg{}, inconsistent cross-slice segmentation, high-frequency artefacts and human mistakes. 
Different from the segmentation masks, the diagnosis labels are independently made by 2 clinicians, and confirmed by 1 senior endocrinologist when diagnoses of the 2 clinicians disagree. We pre-process the dataset from raw 792 CT scans into 1,584 3D image cubes of $128\times 128\times 128$ following Sec.~\ref{sec:distorting_golden_standard}.

As the segmentation mask of adrenal glands is imperfect, we repair the annotation with the proposed NeAR. To demonstrate the usefulness of repairing, we run 3D convolutional networks with the shapes of adrenal glands as inputs, to output the diagnosis labels. The networks are trained with the ALAN dataset, with training / validation / test split of 1,188 / 98 / 298 on a patient level. As will be shown in Sec.~\ref{sec:shape-classification}, the adrenal gland shapes repaired by NeAR could be better classified than the human annotated ones.

The ALAN dataset will be open-source, with 1,584 cases of NeAR-repaired, high-quality 3D models of adrenal glands together with the corresponding diagnosis labels. As there are only a few publicly available medical shape datasets~\cite{yang2020intra,yang2021ribseg} (see supplementary materials for a comparison), our dataset will be a valuable addition to the medical image analysis and geometric deep learning community.

\section{Experiments}

\subsection{Quantitative Experiments on Distorted Golden Standards}
\label{sec:repairing_golden_standards}



\begin{table}[tb]
	\caption{\textbf{Segmentation Repairing on the Distorted Golden Standard Dataset.} We compare the counterparts (Seg-FCN and Seg-UNet), NeAR w/ shape only (S), and NeAR w/ shape and appearance (S+A) in Dice Similarity Coefficient (DSC) and Normalized Surface Dice (NSD) over 5 trials. }
	\label{tab:distorted_golden_standards}
	\centering
    	\begin{tabular}{l|cc|cc}
    		\toprule
    		Metrics & Seg-FCN  & Seg-UNet  & NeAR (S) & NeAR (S+A) \\
    		\midrule
    		DSC (\%, $\uparrow$) & $79.56\pm 0.45$ & $78.70\pm 0.45$ & $78.79\pm 0.45$ & $\mathbf{81.07\pm 0.22}$\\
    		NSD (\%, $\uparrow$) & $89.54\pm 0.33$ & $87.71\pm 0.90$ & $87.96\pm 0.51$ & \bf $\mathbf{91.22\pm 0.12}$ \\
    		%
    		\bottomrule
    	\end{tabular}
\end{table}


\paragraph{Experiment Setting.}

All our experiments are implemented with PyTorch 1.8~\cite{paszke2019pytorch}. To quantitatively analyze the performance of the proposed NeAR method on repairing segmentation annotations, we implement several methods on the distorted golden standard segmentation dataset, including
\begin{itemize}
  \item \textbf{NeAR (S+A)}: The full proposed method;
  \item \textbf{NeAR (S)}: The shape-only NeAR model without appearance $a$ as input;
  \item \textbf{Seg-FCN / Seg-UNet}: Segmentation counterparts, see Sec.~\ref{sec:method-near}.
\end{itemize}
All these methods are trained with all 100 cases, consisting of image and distorted segmentation mask, and evaluated by comparing the similarity between the model-predicted and golden standard segmentation masks. Best models are selected with the lowest training loss. The evaluation is based on volume-based Dice Similarity Coefficient (DSC), and surface-based Normalized Surface Dice (NSD)~\cite{nikolov2018deep} with a distance tolerance of $1.0$.

The segmentation models need to maintain 3D feature maps in the encoder, and thus take much larger memory and computation under the same training resolution as the NeAR. We use a slightly different training schedule. We observed that the number of training iterations of these segmentation models is smaller than the NeAR. Therefore, we utilize an Adam optimizer~\cite{kingma2014adam} with an initial learning rate of $0.001$ for $100$ epochs, delaying the learning rate by $0.1$ after $50$ and $75$ epochs. Longer training schedule does not lead to higher performance.

\paragraph{Results.}


\begin{figure}[tb]
    \centering
	\includegraphics[width=0.8\linewidth]{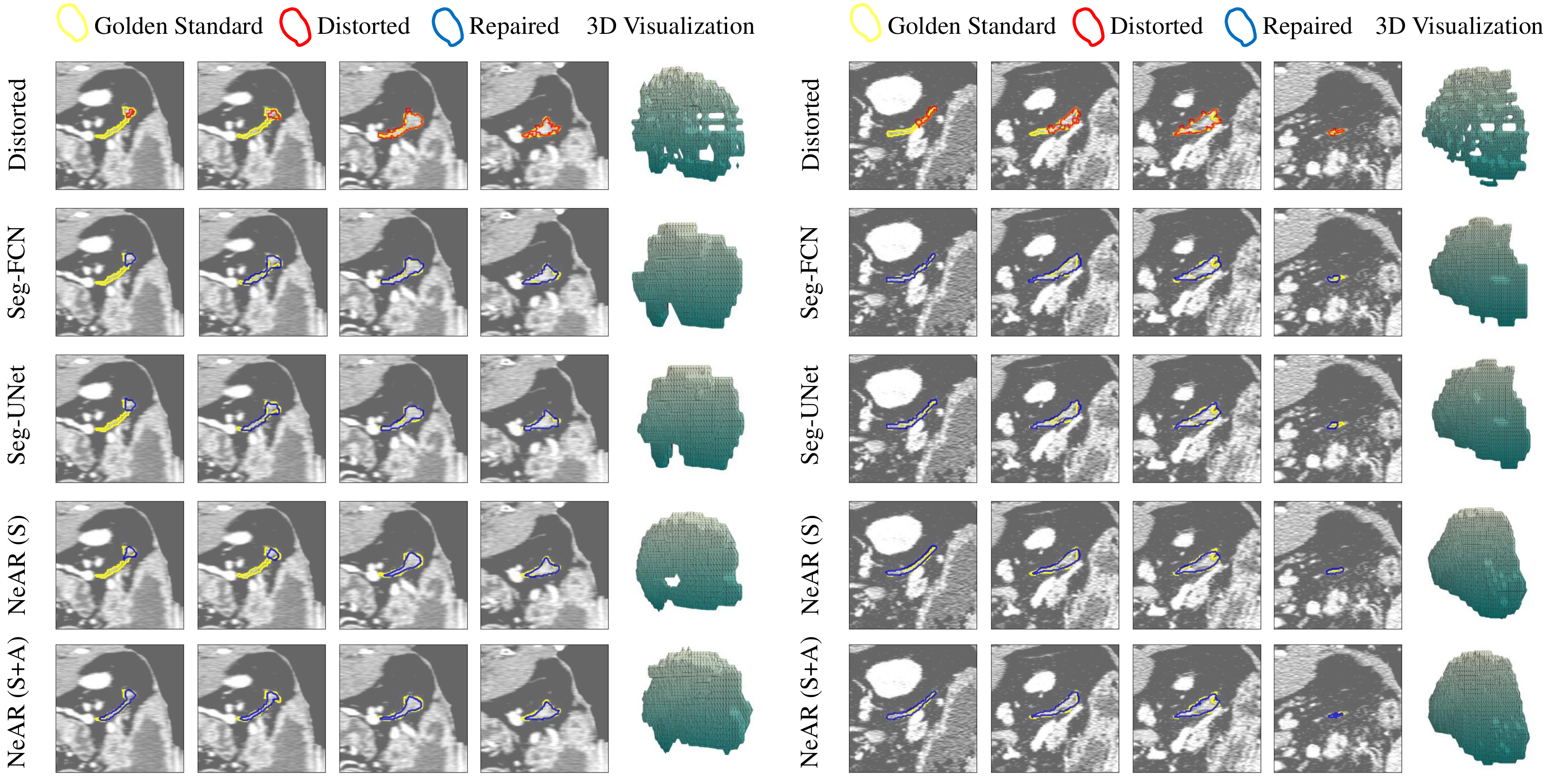}
	\caption{\textbf{Visualization of Repaired Annotations.} Contours of adrenal glands are shown on image slices. Red are distorted, yellow are golden standard, and blue are repaired. The 3D visualization is shown on the right side.}
	\label{fig:visualization_golden_standard}
	\vspace{-15px}
\end{figure}

As depicted in Tab.~\ref{tab:distorted_golden_standards}, the NeAR (S+A) surpasses all the other methods on both DSC and NSD, especially in surface-based method (NSD). NeAR (S) underperforms NeAR (S+A) as well as the standard segmentation method Seg-FCN and Seg-UNet, indicating that appearance-awareness boosts the repair performance significantly. Fig.~\ref{fig:visualization_golden_standard} shows contours of adrenal gland on image slices and 3D visualization of repaired annotations for each method. As shown by the contours on image slices, NeAR (S+A) can repair the distorted annotations more accurately and fix distortions that other methods fail to repair. 

Moreover, we add manual smoothing as a baseline, including morphological closing and connected components filtering. We tried several settings, and the highest Dice is $76.90\%$, much lower than neural methods.

\subsection{Adrenal Diagnosis on the Repaired ALAN Dataset}
\label{sec:shape-classification}

\paragraph{Experiment Setting.}

As described in Sec.~\ref{sec:alan_dataset}, the segmentation masks in the ALAN dataset are imperfect. We utilize 4 different methods to repair the 3D adrenal gland analysis dataset, standard segmentation methods Seg-FCN and Seg-UNet as the baseline methods, NeAR (S) and NeAR (S+A). These methods are conducted in the same settings as that in Sec.~\ref{sec:repairing_golden_standards} respectively. As there are no golden standard segmentation masks in the ALAN dataset, we only provide the qualitative results by visualizing the repaired segmentation masks in the supplementary materials. 

To quantitatively analyze the annotation repairing quality, we conduct shape classification experiments on the human-annotated and model-repaired ALAN dataset. ResNet~\cite{he2016deep} variants with 2D / 3D / ACS~\cite{ACSConv} convolutions are implemented to classify the shapes into binary diagnosis labels (normal vs. abnormal). The shapes of $128\times128\times128$ are resized to the size of $48\times 48\times 48$ as model inputs. For model training, we utilize an Adam optimizer~\cite{kingma2014adam} with an initial learning rate of $0.001$ for $50$ epochs, delaying the learning rate by $0.1$ after $25$ and $40$ epochs. We use cross-entropy loss, and report area under ROC curve (AUC) as the evaluation metric. Best models are selected with lowest validation loss.  We repeat experiments for 5 trials for each setting.

\paragraph{Results.}
As depicted in Tab.~\ref{tab:shape_classification},
models trained with shapes repaired by the NeAR (S+A) can diagnose the adrenal glands better than other methods, as well as the human annotated imperfect ones. Notably, standard segmentation methods (Seg-FCN and Seg-UNet) deliver worse shape classification results than NeAR (S) and raw human annotation, though the segmentation methods produce better shape repairing results than NeAR (S) in Sec.~\ref{sec:repairing_golden_standards}. This implies that the learned prior of deep implicit surfaces can be particularly useful for downstream applications.


\begin{table}[tb]
\scriptsize
	\caption{\textbf{Shape Classification on the ALAN Dataset.} We repair the 3D shapes of adrenal glands using standard segmentation (Seg-FCN and Seg-UNet), NeAR w/ shape only (S), and NeAR w/ shape and appearance (S+A). ResNet-18 and ResNet-50 variants are trained to classify the 3D shapes of adrenal glands (normal vs. abnormal) on the human-annotated and the repaired datasets. We report mean and standard deviation of AUC (\%, $\uparrow$) on the test set over 5 trials.}\label{tab:shape_classification}
	\centering
	\begin{tabular*}{\hsize}{@{}@{\extracolsep{\fill}}l|c|cc|cc@{}}
		\toprule
		Networks & Human-Annotated & Seg-FCN & Seg-UNet & NeAR (S) & NeAR (S+A)  \\
		\midrule
		ResNet-18~\cite{he2016deep} (2D) & $68.35\pm 2.53$ & $65.17\pm 2.21$ & $64.68\pm 2.45$ & $65.95\pm 0.83$ & $\mathbf{69.69 \pm 1.44}$\\
		ResNet-18~\cite{he2016deep} (3D) & $89.77\pm 1.20$ & $86.27\pm 1.91$ & $86.55\pm 0.89$ & $88.64\pm 1.24$ & $\mathbf{90.38 \pm 0.57}$\\
		ResNet-18~\cite{he2016deep} (ACS~\cite{ACSConv}) & $90.10 \pm0.90$ & $87.02\pm 2.20$ & $86.83\pm 2.00$ & $89.22\pm 1.08$ & $\mathbf{91.11 \pm 0.35}$\\
		\midrule
		ResNet-50~\cite{he2016deep} (2D) & $66.36\pm 2.56 $ & $64.88\pm 3.47$ & $66.06\pm 2.79$ & $69.04\pm 3.11$ & $\mathbf{69.94\pm 1.18}$\\
		ResNet-50~\cite{he2016deep} (3D) & $89.72\pm 1.33$ & $85.64\pm 1.59$ & $84.92\pm 1.08$ & $88.76\pm 0.91$ & $\mathbf{89.78\pm 0.79}$\\
		ResNet-50~\cite{he2016deep} (ACS~\cite{ACSConv}) & $90.13\pm 0.40$ & $85.91\pm 2.11$ & $82.28\pm 2.34$ & $89.42\pm 1.35$  & $\mathbf{90.72\pm 0.72}$\\
		\bottomrule
	\end{tabular*}
	\vspace{-10px}
\end{table}


\section{Conclusion}
This study addresses a practical problem in medical image analysis: how to repair the imperfect segmentation. We propose Neural Annotation Refinement, an appearance-aware implicit method, whose values are validated in repairing segmentation and downstream applications. Moreover, the ALAN dataset for 3D shape classification will be an addition for the research community. There are limitations in the current study, \eg{}, validated on adrenal glands only. We will test the NeAR on sparse annotations and small objects in the future research.

\subsubsection{Acknowledgment.}
This work was supported by National Science Foundation of China (U20B2072, 61976137), supported in part by a Swiss National Science Foundation grant, and also supported in part by Grant YG2021ZD18 from Shanghai Jiao Tong University Medical Engineering Cross Research.

\bibliographystyle{splncs04}
\bibliography{string,reference}

\clearpage

\appendix

\setcounter{table}{0}
\renewcommand{\thetable}{A\arabic{table}}

\setcounter{figure}{0}
\renewcommand{\thefigure}{A\arabic{figure}}


\begin{table}[!htb]
	\caption{\textbf{A Comparison of Public Available Medical Shape Datasets.}}
	\label{tab:medical_shape_datasets}
	\centering
	    \begin{tabular*}{\hsize}{@{}@{\extracolsep{\fill}}lccc@{}}
    		\toprule
    		 & IntrA~\cite{yang2020intra} & RibSeg~\cite{yang2021ribseg} & ALAN \\
    		\midrule
    		Task & Classification \& Segmentation & Segmentation & Classification \\
    		Body Part & Vessel & Rib & Adrenal Gland  \\
    		Characteristics & Tubular & Tubular & Soft Organ\\
    		No. Cases & 1,909 & 490 & 1,584\\
    		\bottomrule
    	\end{tabular*}
\end{table}


\begin{figure}[!htb]
    \centering
	\includegraphics[width=0.8\linewidth]{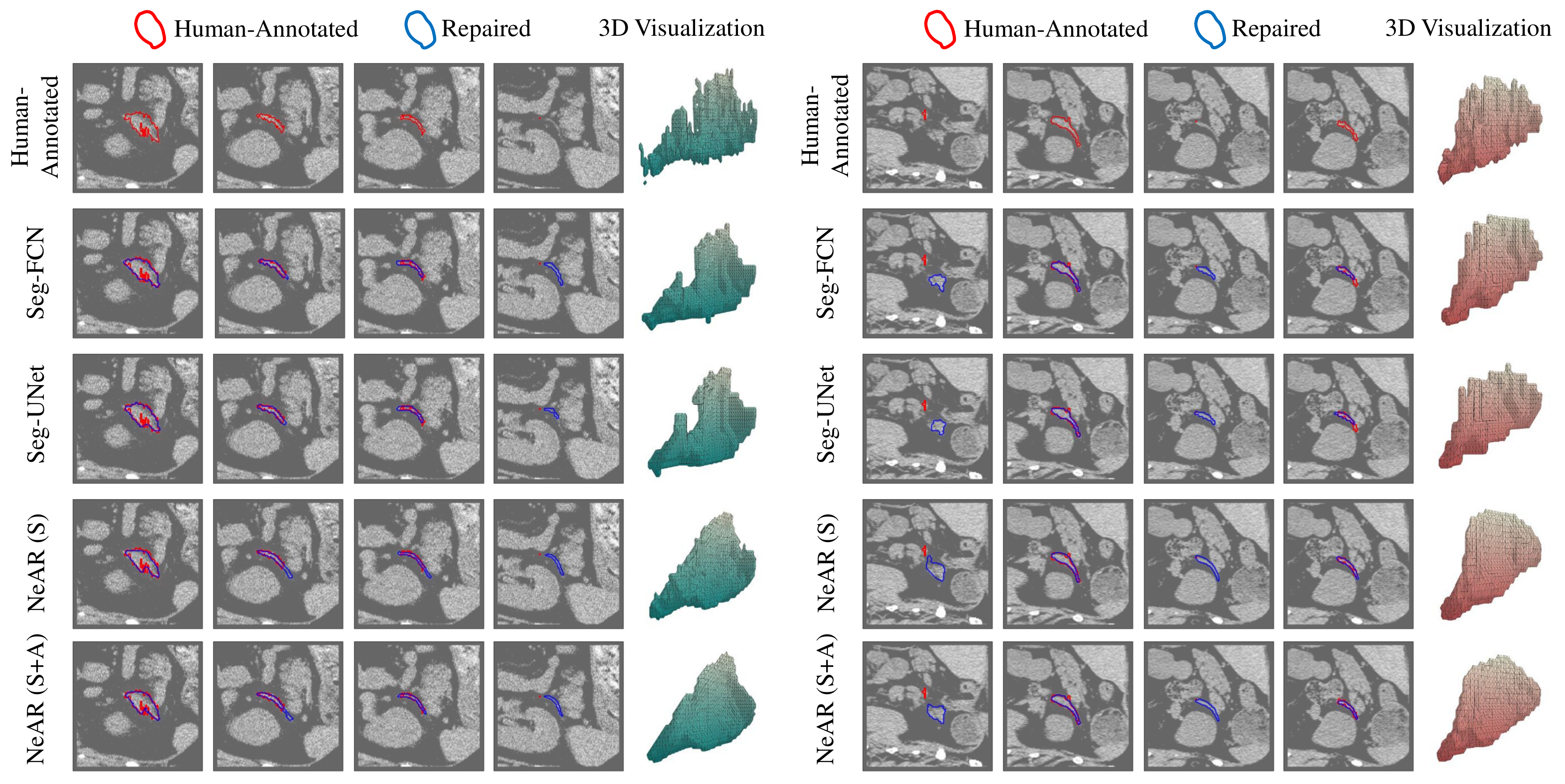}
	\caption{\textbf{Visualization of Two Samples in the Repaired ALAN Dataset.} Contours of adrenal glands are shown on image slices. Red are human-annotated, and blue are repaired. The 3D visualization is shown on the right side, green/red denotes normal/abnormal adrenal glands.}
	\label{fig:visualization_alan}
\end{figure}

\end{document}